\documentclass[letterpaper, 10 pt, conference]{ieeeconf}  % Comment this line out if you need a4paper

\IEEEoverridecommandlockouts                              % This command is only needed if 
\usepackage{graphicx} % for pdf, bitmapped graphics files
\usepackage{color}
\usepackage{acronym}
\usepackage{epsfig} % for postscript graphics files
\usepackage{amsmath} % assumes amsmath package installed
\usepackage{amssymb}  % assumes amsmath package installed
\usepackage{hyperref}

\newcommand{\secref}[1]{Section~\ref{#1}}

\newcommand{\figref}[1]{Figure~\ref{#1}}

\title{\LARGE \bf
Learning Camera Miscalibration Detection
}

\author{Andrei Cramariuc*, Aleksandar Petrov*, Rohit Suri, Mayank Mittal, Roland Siegwart, and Cesar Cadena% <-this % stops a space
\thanks{All the authors are with the Autonomous Systems Lab, ETH, Zurich 8092, Switzerland, {\tt\small \{crandrei, alpetrov, suriro, mittalma, rsiegwart, cesarc\}@ethz.ch}}%
\thanks{* marks equal contribution}
}

\begin{document}

\maketitle
\thispagestyle{empty}
\pagestyle{empty}

%%%%%%%%%%%%%%%%%%%%%%%%%%%%%%%%%%%%%%%%%%%%%%%%%%%%%%%%%%%%%%%%%%%%%%%%%%%%%%%%

\begin{abstract}

Self-diagnosis and self-repair are some of the key challenges in deploying robotic platforms for long-term real-world applications.
One of the issues that can occur to a robot is miscalibration of its sensors due to aging, environmental transients, or external disturbances. 
Precise calibration lies at the core of a variety of applications, due to the need to accurately perceive the world.
However, while a lot of work has focused on calibrating the sensors, not much has been done towards identifying when a sensor needs to be recalibrated.
This paper focuses on a data-driven approach to learn the detection of miscalibration in vision sensors, specifically RGB cameras. 
Our contributions include a proposed miscalibration metric for RGB cameras and a novel semi-synthetic dataset generation pipeline based on this metric.
Additionally, by training a deep convolutional neural network, we demonstrate the effectiveness of our pipeline to identify whether a recalibration of the camera's intrinsic parameters is required or not.
The code is available at \url{http://github.com/ethz-asl/camera_miscalib_detection}.

\end{abstract}

%%%%%%%%%%%%%%%%%%%%%%%%%%%%%%%%%%%%%%%%%%%%%%%%%%%%%%%%%%%%%%%%%%%%%%%%%%%%%%%%

\section{Introduction}
\label{sec:introduction}

In robotics, errors in the estimation of the system's parameters can adversely affect the accuracy of algorithms for state estimation and the performance of feedback controllers.
In order to avoid systematic errors due to incorrect parameter estimates,  a common practice is to perform sophisticated calibration of the system by a human expert~\cite{roth1987calibration}.
Once determined, calibration parameters are kept fixed during the operation cycle of the robot.
However, this approach is not sustainable for a variety of real-world applications where robots need to operate in harsh environments for extended periods of time.
The calibration parameters of the system are prone to change over time due to component wear, environmental transients such as temperature changes, or external disturbances like collisions.
Additionally, it may be impractical to perform offline calibration regularly as a means to address this issue.

An alternative solution to offline calibration is online calibration techniques that are performed during the system's normal operation, such as those presented in~\cite{preiss2018simultaneous,schneider2017viso}.
These techniques, though promising, are computationally expensive and have various limiting requirements, such as the type of required motion or the storage and processing of data to create a calibration dataset.
Hence, instead of running these methods periodically and recalibrating the robotic platform, ideally, one would like to perform the calibration only when the system is detected to be miscalibrated.
This objective is considered as a constituent of the fault detection and diagnosis for a robotic system~\cite{visinsky1994robot}.

Since sensors lie at the core of any autonomous system, it is critical to detect the sensor data faults for safety and stable performance~\cite{ni2009sensor}.
However, unlike sensors for measuring attitude or temperature, calibration errors in vision sensors do not appear as an offset or as a drift in the sensor's readings.
Although having hardware redundancy is a way to detect imperfections, it increases the cost and complexity of the system.
Further, due to their complex nature, it is difficult to obtain a unified analytical solution for identifying miscalibration in vision sensors.
Fortunately, common operating environments, both indoors and outdoors, contain regularities that can be exploited for this purpose, such as walls, furniture, street lamps, etc. 
%
%We conjecture that these regularities are not robust to the perturbations that arise when observed by a disturbed camera setup. 
%
%For instance, under incorrect distortion parameters, a straight pole in the incorrectly rectified camera image may appear curved.
%
%With the recent success of deep learning, it can be reasonably expected that a neural network should be sensitive to these perturbations and help in camera miscalibration detection.
%
We propose a data-driven approach that implicitly utilizes these regularities.
%
%We demonstrate that this is indeed the case.

\begin{figure}[!t]
\centering
\vspace{4pt}
\includegraphics[width=.98\linewidth]{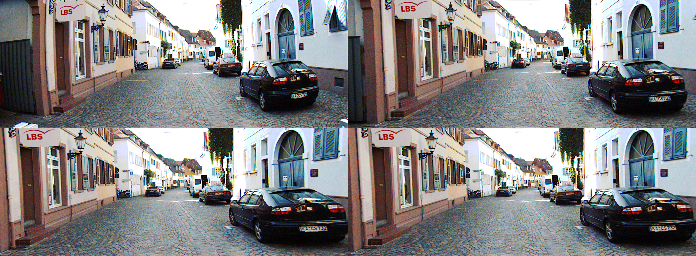}
\caption{
Illustration of the subtle differences that a miscalibration detection system needs to be sensitive to.
\textbf{Top left:} Unrectified image;
\textbf{Top right:} Correctly rectified image;
\textbf{Bottom:} Two examples of incorrectly rectified images. The image is taken from the KITTI dataset~\cite{geiger2013vision}.
}
\label{fig:teaser-image}
\vspace{-12pt}
\end{figure}

In this work, our goal is not to provide a neural network that detects miscalibrations for any camera. 
%
%Such a level of generalization is difficult to obtain.
%
Instead, we propose a method in which a network is tuned for a specific camera to predict when an automatic recalibration is necessary for that camera.
However, using a learning-based approach poses its own set of challenges.
First, a large-scale dataset for training a network to detect miscalibration is not currently available in the public domain.
Second, there is no standard metric for measuring the degree of miscalibration in the intrinsics of a camera.
We address these challenges and provide the following key contributions: 
\begin{itemize} 
    \item A novel dataset generation pipeline to create a large-scale dataset for camera miscalibration detection. 
    \item A metric, \textit{average pixel position difference}, for estimating the degree of miscalibration and analysis of how it correlates with performance in a monocular odometry task.
    \item A deep convolutional neural network (CNN) that predicts when the camera is miscalibrated even in previously unseen scenes.
\end{itemize}

\section{Related Work}
\label{sec:related}

In the last decade, a variety of approaches have been proposed for calibration of various types of range-based sensors, inertial sensors, and vision sensors, as well as the extrinsic calibration between them.
In this section, we focus on literature related to sensor miscalibration, the estimation of intrinsic parameters of vision sensors, and fault detection in multi-sensor systems.

Accurate camera calibration is an important step for a multitude of 3D computer vision tasks.
The existing calibration techniques can be broadly categorized as photogrammetric calibration and self-calibration.
In photogrammetric calibration, the camera calibration is performed by observing a target of known geometry in 3D space.
Over the years, various types of tags such as checkerboards~\cite{zhang2000calib, scaramuzza2006toolbox} and fiducial markers~\cite{atcheson2010caltag} have been proposed for this purpose.
These approaches typically pose the calibration problem as a non-linear optimization problem to minimize a reprojection error and to estimate the most likely values of the camera parameters.
However, the need to have an apparatus and a human expert in these techniques prevents them from being scalable or practical for robots deployed into the real world.

On the other hand, self-calibration, as introduced in~\cite{maybank1992theory}, does not require a calibration object.
Through a sequence of images, these methods estimate the intrinsic parameters that are consistent with the underlying projective reconstruction of the observed scene.
Certain approaches use camera motion constraints, such as planar motion~\cite{armstrong1996self} or rotation of the camera~\cite{agapito2001self}, in conjunction with the 3D metric reconstruction of the scene to calibrate the camera's intrinsic parameters.
Sturm~\cite{sturm1997critical} presents the concept of critical motion sequences for a camera with constant parameters for which there exists no unique solution for self-calibration.
Wildenauer and Hanbury~\cite{wildenauer2012robust} detect orthogonal vanishing points in the scene to generate a hypothesis for focal length.
However, the flexibility provided by self-calibration techniques comes at the price of computation expenses. 

More recently, with the advent of deep learning~\cite{Goodfellow2016DL}, data-driven approaches have also been proposed to estimate the calibration parameters of the camera.
Workman \textit{et al.}~\cite{workman2015deepfocal} propose a CNN for estimating the focal length of an image.
To train the network, they construct a dataset by combining images and camera models estimated using 1D structure from motion~\cite{wilson2014robust}.
On the other hand, Lopez \textit{et al.}~\cite{lopez2019deep} use separate regressors, which share a common pre-trained network architecture, to estimate tilt, roll, focal length, and radial distortion parameters from a single image.
They use the SUN360 panorama dataset~\cite{xiao2012recognizing} to artificially generate the training images. 
However, the estimation of these parameters is highly dependent on finding the horizon in the image, an assumption that is highly environment dependent.
Unlike the previous two approaches, which aim to estimate the camera calibration parameters, Yin \textit{et al.}~\cite{yin2018fisheyerecnet} propose an end-to-end multi-context deep network for removing distortions from single fish-eye camera images.
They use a scene-parsing network to provide semantic cues during training and use an $L_2$ reconstruction loss for rectified image prediction.

Fault detection in multi-sensor systems can be done by correlating the information from multiple sensors and rejecting measurements that do not match~\cite{mendoza2012mobile,sundvall2006fault,roumeliotis1998sensor,lu2004parity,schneider2019observability}.
These methods are generally able to detect when a fault has occurred, however they rely on a redundant sensor setup.
When the fault estimation is done indirectly, through an intermediary task such as localization performance, it can be ambiguous to decide whether the sensor is at fault or if the localization system failed.

While some of the above mentioned approaches deal with estimating the calibration parameters through either a geometry-based or a learning-based approach, our work is orthogonal and does not aim to replace them.
We want to complement these methods by identifying when a camera needs to be recalibrated.
Further, we do not want to rely on sensor redundancy since that increases the cost of the system.
Thus, our objective is to detect miscalibration in a single camera.
To the best of our knowledge, this is the first work proposing a deep learning approach to detect miscalibration of the intrinsic parameters for an RGB camera.

\section{Methodology}
\label{sec:method}
In this section, we present our contributions in detail.
The dataset generation pipeline is explained in \mbox{\secref{subsec:dataset-generation}}.
In \secref{subsec:dataset-metric}, our novel metric for miscalibration is defined.
The neural network and its training are detailed in \secref{subsec:network}.

\subsection{Dataset Generation}
\label{subsec:dataset-generation}

A straightforward procedure to create a dataset for camera miscalibration detection is to manually vary the camera parameters by using different lenses while taking any one of the settings as the nominal one.
However, this process is time-consuming and tedious since offline calibration would be required for every new setting.
The procedure is also limited with respect to the generation of disturbances in the camera parameters. 
Some cameras have only one degree of freedom for calibration (the distance between the lens and the sensor), hence the calibration parameters cannot be varied independently.
%
%In addition, as the network must be retrained for each camera, this method of collecting separate datasets is impractical.
%
Due to these limitations, we propose an alternative solution to generate a semi-synthetic dataset by using a set of raw images and a set of correct calibration parameters for a given camera setup.
The presented method is based on the idea that the visual effect obtained from rectifying an image from a miscalibrated sensor with its initial belief of the parameters is similar to the effect of rectifying an image from a calibrated sensor with parameters different from the correct ones.

In our semi-synthetic dataset generation pipeline, we consider the pinhole camera model with radial and tangential distortion~\cite{heikkila2000geometric}.
We denote the set of true calibration parameters of the camera model as $\Theta = \{f_u, f_v, u_c, v_c, \mathbf{k}_r, \mathbf{k}_t\}$.
Consider the raw camera image $I$, which is rectified using the parameters $\Theta$ to obtain the rectified image $I'$.
The rectification map $M' = f(\Theta)$ used in this process relates each pixel in the rectified image to a position in the original image.
%
%We denote the image resulting from applying $M'_\text{correct}$ to $I$ as $\hat{I}_\text{correct}'$.
%
Generally, not all the pixels in the rectified image have a corresponding position in the original one. 
Therefore, we define a validity mask, $R$, which is the largest rectangular region in the rectified image $I'$ with only valid pixels, and that has the same aspect ratio as the original image $I$.
The final sample image $\hat{I}$ is obtained by first cropping the valid mask region $R$ of the image $I'$ and then rescaling the result to the size of the original raw image $I$.
An alternate way to express the rectification is by applying the rectification map $\hat{M}$ (which is obtained by cropping and rescaling $M'$) on the raw image $I$ to directly obtain the final sample image $\hat{I}$.
%
%Since this process only depends on the considered calibration parameters and not on the raw image, we can write the entire rectification process as, $\hat{I} = h(I; \Theta)$.

In general, it is difficult to obtain the true calibration parameters of the camera. 
Thus, we use the values estimated using a calibration toolbox as the correct calibration parameters $\Theta^\star$ and denote the correct rectified image and rectification map as $\hat{I}^\star$ and $\hat{M}^\star$ respectively. 
To obtain samples of miscalibrated images, we perturb each intrinsic parameter independently to obtain $\Theta^m$. 
This process allows generating arbitrarily many miscalibrated images, $\hat{I}^m$, and rectification maps, $\hat{M}^m$, by randomly perturbing the parameters.
Thus, by collecting only a set of raw images with a correct calibration of the sensor, one can generate a large amount of data for detecting camera miscalibration.
Even though we consider a pinhole camera model, this approach is also applicable to other camera models.

%\begin{table}[t]
%\caption{Sampling range $\Delta$ expressed as percentage relative change %of correct camera parameters $\hat{\Theta}$}
%\label{tab:sampling-settings}
%\centering
%\begin{tabular}{c c c}
%    \hline 
%    Parameter & Min. Value & Max. Value \\
%    \hline
%    $f_u, f_v$ & $-5\%$  & $20\%$\\
%    $c_u, c_v$ & $-5\%$ & $5\%$\\
%    $k_1, k_2, k_3, p_1, p_2$ & $-15\%$ & $15\%$ \\
%    \hline
%\end{tabular}
%\end{table}

\begin{figure}[!t]
\centering
\vspace{6pt}
\includegraphics[width=.9\linewidth]{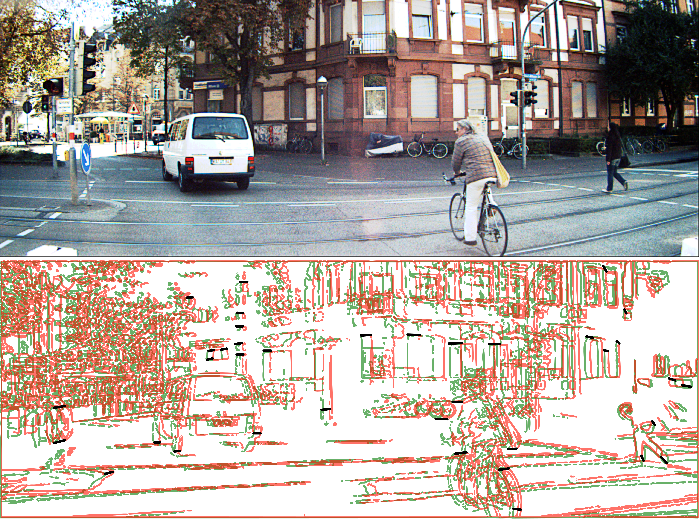}
\caption{
\textbf{Top:} An unrectified image from the KITTI dataset~\cite{geiger2013vision}.
\textbf{Bottom:} For illustration purposes, canny edges detected from correctly (in green) and incorrectly (in red) rectified images are shown.
A set of incorrect rectification parameters results in pixel projections from the raw image to be displaced relative to that with the correct parameters (indicated by the black segments).
The mean of the L2-norms of these displacements over the image corresponds to the APPD.}
\label{fig:disturbed-rectification}
\end{figure}

\begin{figure}[!t]
\centering
\includegraphics[width=0.9\linewidth]{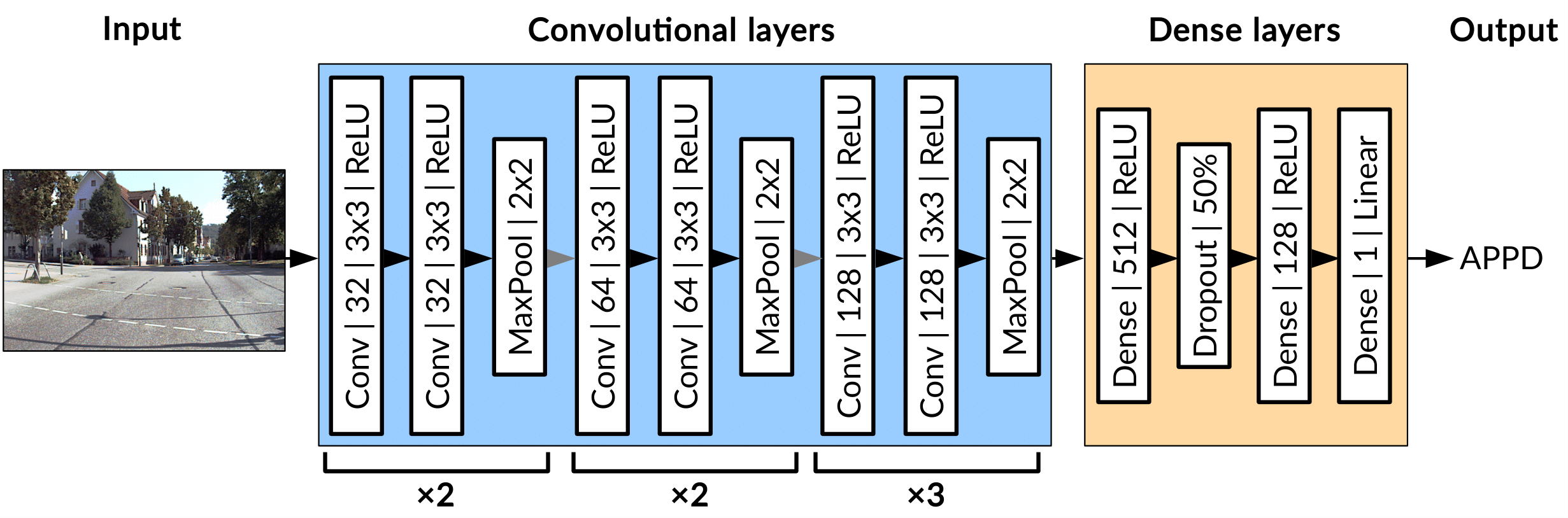}
\caption{
The network architecture used to run the experiments.
%
%As parameters, the convolutional layers have the number of filters and their sizes, the max pool layers have the size of the pooling operation, the dense layers have the number of nodes, and the dropout layers have the percentage of dropped values.
%
All layers except the last one use Rectified Linear Unit (ReLU) activation functions.
}
\label{fig:miscalib-network}
\end{figure}

\subsection{Metric for Degree of Miscalibration}
\label{subsec:dataset-metric}

As described in~\secref{subsec:dataset-generation}, image rectification is a transformation parameterized by the calibration parameters.
Since these parameters are continuous, one can generate images arbitrarily close to a correctly rectified image by applying small perturbations to the true calibration parameters.
Due to the non-linear effects and the strong correlations of these parameters on the rectification transformation, defining a meaningful distance metric to directly assess the quality of different randomly chosen calibrations is difficult.
Moreover, as the rectification of an input image is typically only the first stage of a system, the degree of miscalibration should be considered in conjunction with the corresponding reduction in the overall system performance.
Therefore, we propose an indirect approach using the \textit{average pixel position difference (APPD)} as a scalar metric to measure the degree of camera miscalibration.
%
%As an example, we show in \secref{subsec:appd_and_reprojection} how the APPD metric directly correlates with the inaccuracies caused by an incorrect calibration to a typical vision-based SLAM system.

Using the symbols introduced in~\secref{subsec:dataset-generation}, the numeric value for the APPD, denoted by $\delta$, is calculated using the rectification maps $\hat{M}^\star$ and $\hat{M}^m$ obtained from using calibration parameters $\Theta^\star$ and $\Theta^m$ respectively.
Since these maps are computed using different parameters, they relate the same pixel coordinate in their corresponding rectified images to a different image coordinate in the raw image.
The Euclidean distance between these two positions is referred to as the \textit{pixel position difference}.
This is illustrated in \figref{fig:disturbed-rectification}.
The APPD is the mean value of these pixel position differences over the entire image, i.e.
\begin{equation*}
    \delta = \frac{1}{H \times W} \sum_{p \in I}||\hat{M}^\star(p) - \hat{M}^m(p)||_2,
\end{equation*}
where $(H, W)$ is the size of the image $I$ and $p$ denotes the pixel coordinate $(u, v)$.
Even when normalized by the number of pixels, the value still depends on the resolution.
Normalizing further by the diagonal makes it resolution-independent.
That is why we report APPD values as a percentage of the image diagonal, i.e. they are divided by the diagonal and scaled by a factor of 100.
%
%A deeper analysis in the relationship between the APPD and the reprojection error is presented in \secref{subsec:appd_and_reprojection}.

\begin{figure*}[!t]
\centering
\includegraphics[width=0.90\linewidth]{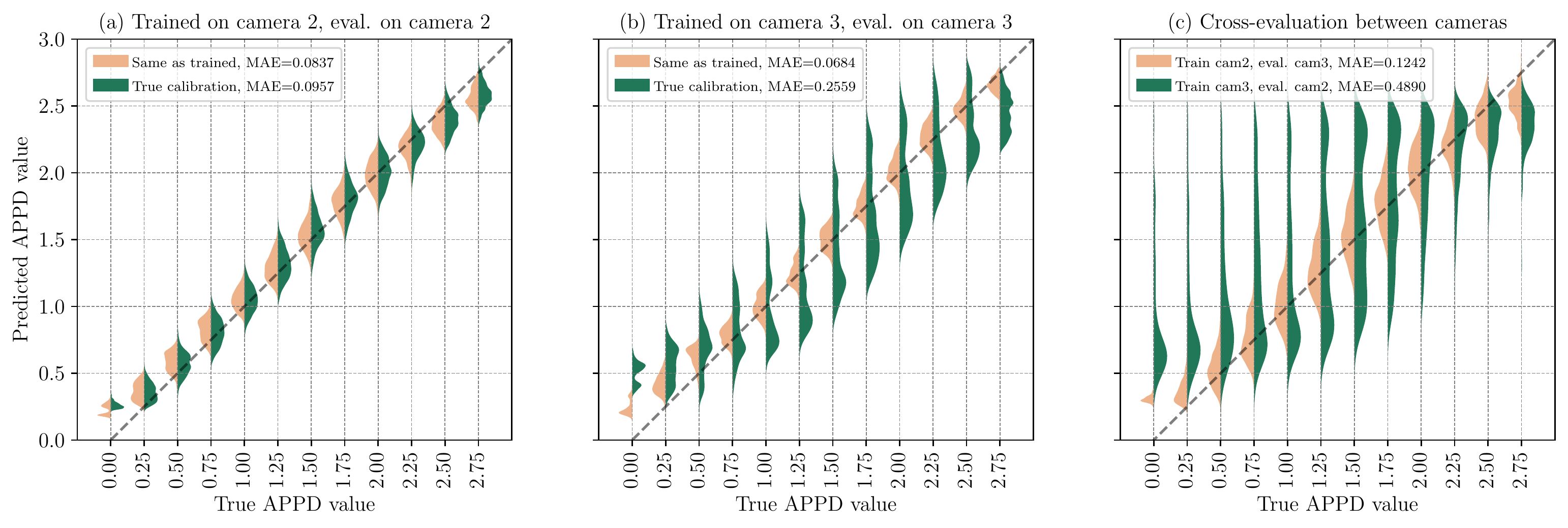}
\caption{APPD prediction accuracy of the trained neural network models for the two RGB cameras from KITTI~\cite{geiger2013vision}.
The plots show the distributions of networks predictions for given quantized APPD values.
The dashed line designates perfect prediction.
}
\label{fig:appd_pred}
\end{figure*}

\subsection{Network Architecture and Training}
\label{subsec:network}

The architecture of the APPD prediction network is presented in~\figref{fig:miscalib-network}. 
The input to the network is the rectified image $\hat{I}^m$, and the output is the APPD metric.
To prevent artifacts and loss of minute details due to image resizing, we use the input at full resolution. 

For each camera and corresponding correct calibration, we train a separate network to deploy alongside the respective camera.
This can be seen as an addition to the calibration procedure that, in a similar manner, is also pre-computed for each camera separately.
The goal of the dataset generation process described in~\secref{subsec:dataset-generation} is to reduce the amount and variety of data required to train the model.
With the proposed method, it is sufficient to collect a single dataset, with correct calibration known and without any manual labeling.

During training, the perturbed parameters are sampled such that the calculated APPD values follow an approximately uniform distribution.
Additionally, 1\% of the samples are kept with the correct rectification, i.e. APPD value of zero.
We use a mean squared error loss between the network predictions and the ground-truth labels for training. 
This loss is optimized by using the Adaptive Moment Estimation (ADAM) optimizer~\cite{adam}.
We initialize the network parameters by Xavier's initialization method~\cite{glorot2010understanding} and use dropout~\cite{srivastava14dropout} to avoid overfitting.

\section{Experiments and Discussion}

The results from evaluating the trained models are discussed in~\secref{subsec:miscalib_detection_nn}. 
We present some generalization results for our approach in~\secref{subsec:generalization}.
The relationship between APPD and the intrinsic parameters is experimentally investigated in~\secref{subsec:appd_and_parameters}.
\secref{subsec:appd_and_reprojection} compares APPD and reprojection error.

\begin{figure*}[!t]
\centering
\includegraphics[width=0.65\linewidth]{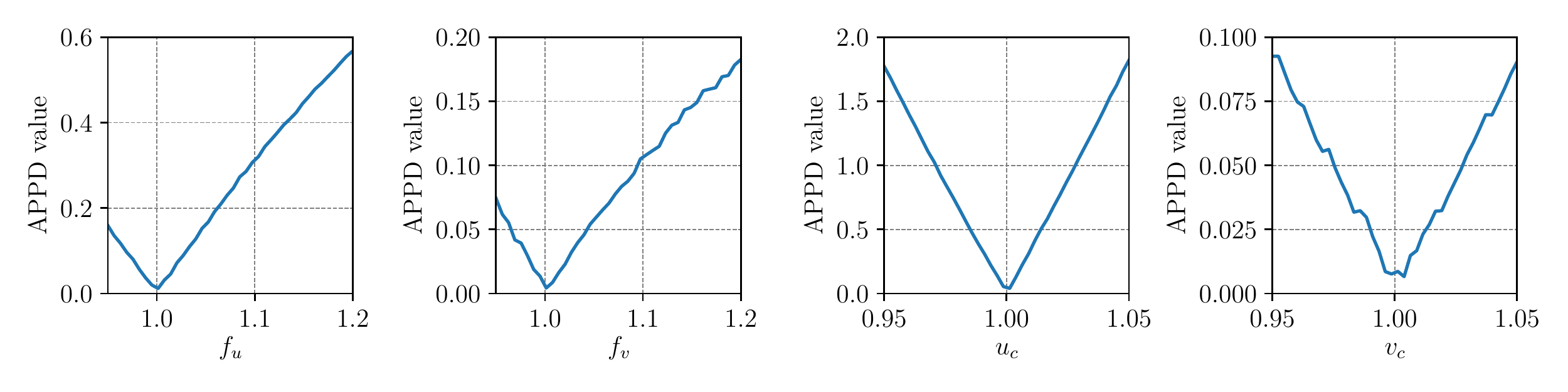}
\includegraphics[width=0.82\linewidth]{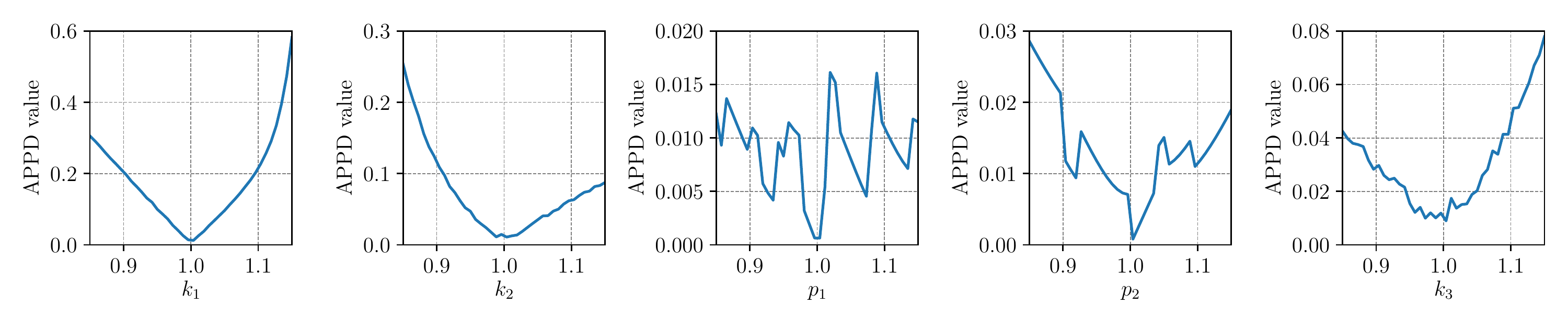}
\caption{
Plots showing the different effects of the camera's intrinsic parameters on the APPD, when one parameter is varied, and the rest are kept fixed.
The x-axis is the multiplication factor applied to the reference parameter.
The used reference calibration is from camera 2 for the KITTI sequences from date $26.09.2011$.
Note that the y-axis scales in the plots for $f_u$ and $f_v$, as well as for $u_c$ and $v_c$, are different.
This is due to the image's aspect ratio.
}
\label{fig:params-to-appd}
\end{figure*}

\subsection{Detection of Miscalibrations with a Neural Network}
\label{subsec:miscalib_detection_nn}

The KITTI dataset has two RGB cameras (\mbox{cameras 2 and 3}).
We split the KITTI sequences from September 26, 2011 to obtain our training and validation sets.
For testing, we use all sequences from the other four days.
We vary the focal lengths from $-5\%$ to $20\%$, the optical center $\pm 5\%$, and the distortion coefficients $\pm 15\%$.
The dataset provides different calibration files for every day, which were observed to be inconsistent.  
It is not known whether the cameras differed physically on different days, or if the differences in the calibrations arise from imperfections in the calibration procedure.
Therefore, there is no single `correct' calibration that can be used as a reference for calculating the true APPD value when evaluating prediction performance.
Instead, we consider two cases: (i) taking the set of parameters corresponding to the day used for training, and (ii) using the set corresponding to the day on which the test image was actually recorded.

\mbox{Figures \ref{fig:appd_pred}a and \ref{fig:appd_pred}b} show the prediction quality of the trained networks for both camera 2 and camera 3, evaluated for the two cases described above.
The mean absolute error (MAE) for each case is also reported.
It can be seen that the models are able to generalize also to images and environments they have not seen before.
While both networks are powerful in detecting miscalibration with respect to the reference set of parameters that they were trained with, the one for camera 2 performs significantly better.

This mismatch in performance is caused by the level of similarity between the sets of correct calibration parameters provided for each camera.
The APPD ranges for the four test days, relative to the day used for training and validation are $[0.12, 0.93]$ and $[0.78, 2.37]$ for camera 2 and camera 3 respectively. 
It is likely that camera 3 might not have been well-calibrated either on the training day or on some of the other days.
This result illustrates the importance of selecting a `correct' calibration, with respect to which the training process must be defined.

As the two cameras are of the same make and brand, are positioned solely with a horizontal offset from one another, and operate in the same environment, the transferability of the model trained on the data from one camera to the other was also evaluated.
The corresponding results are shown in~\mbox{\figref{fig:appd_pred}c}.
Indeed, the model trained on \mbox{camera 2} generalizes well to camera 3.
\mbox{\figref{fig:appd_pred}c} also shows that the reverse generalization does not hold, which stresses the importance of the choice of reference calibration and is another indication that camera 3 might have been slightly less consistently calibrated.

\figref{fig:appd_pred} further demonstrates that the trained models experience bias in the extremely low and extremely high APPD values.
This is a limitation of both training in a regression setting and of the miscalibration sampling procedure, which provides very few miscalibrations with APPD values close to 0.
Instead, if one targets a specific performance metric, which can be related to APPD (see~\secref{subsec:appd_and_reprojection}), then they can determine a threshold value and rephrase the problem into a binary classification setting.

While the presented neural network architecture is simple and further performance improvement may be possible, the above results indicate that a CNN can indeed be trained to be sensitive to miscalibration artifacts.
One should note that the data does not explicitly designate the regularities which are not robust to the perturbation effects arising from disturbing a camera setup, but the model has discovered these regularities on its own.
Moreover, even though motion distortion and blur are not explicitly addressed by the analysis, they are represented in both the training and test sets (as the images are obtained from a moving vehicle), and therefore the results account for them as well.

\subsection{Generalization to new Environments and Cameras}
\label{subsec:generalization}

\begin{figure}[!t]
\centering
\includegraphics[width=1.0\linewidth]{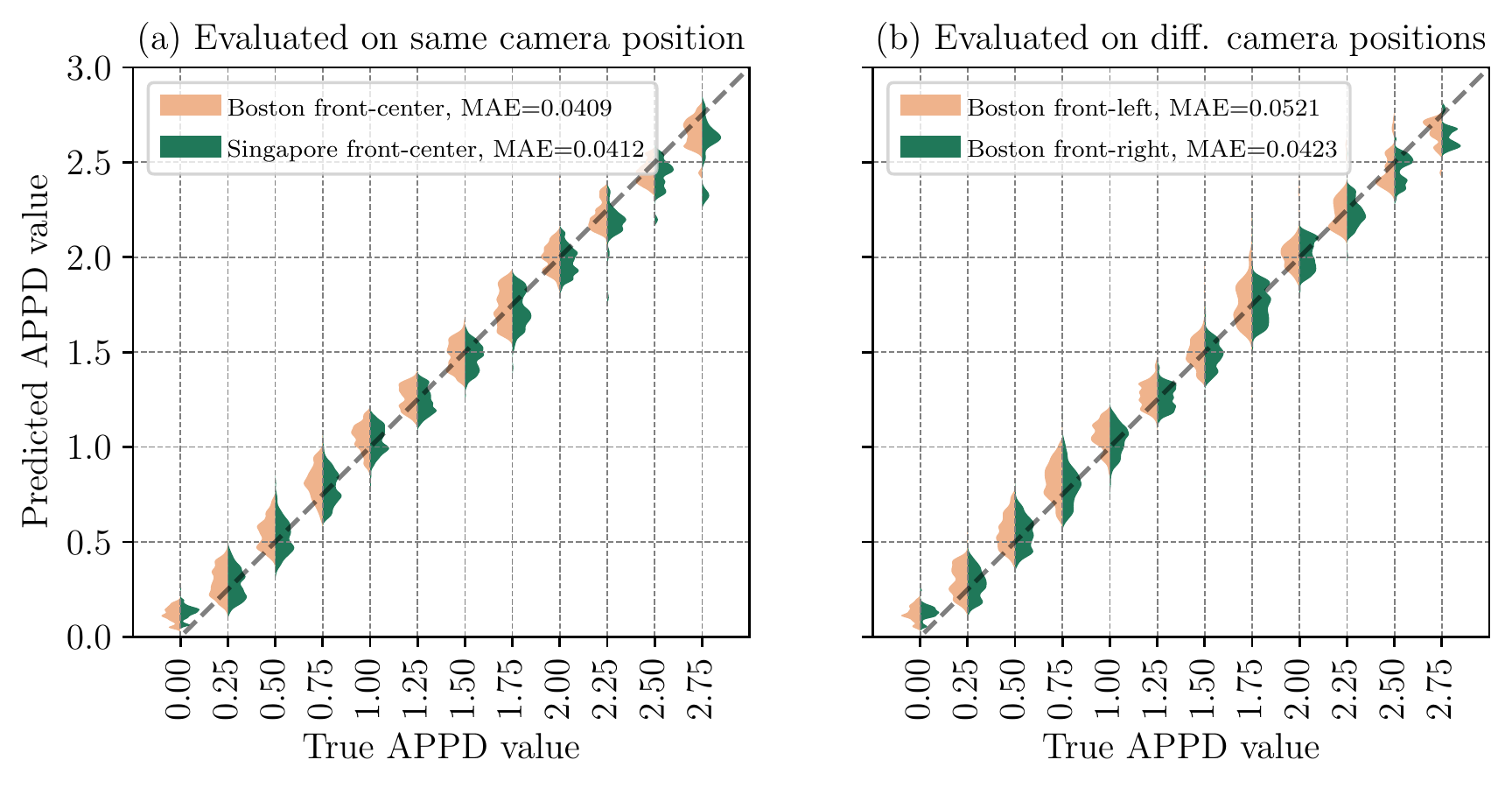}
\caption{APPD prediction accuracy when evaluating on environments and camera positions that were not represented in the training set.
The training was performed on a subset of the nuScenes dataset \cite{nuscenes2019} recorded with the front-center camera in Boston.
The plots show the distribution of predictions for given quantized APPD values.
The dashed line designates perfect prediction.}
\label{fig:appd_pred_nuscenes}
\vspace{-6pt}
\end{figure}

The KITTI dataset is limited in the variation of its scenes (recordings only in the city of Karlsruhe, Germany), and in the position of the camera sensors (both oriented forward with only a horizontal offset between them).
In order to study the potential further generalization capabilities of the proposed method, we trained the same model on some of the scenes recorded in Boston from the forward camera of the nuScenes dataset~\cite{nuscenes2019}. 
Only the scenes recorded during the day were considered.

The performance of the model was evaluated on the other scenes of the same camera in Boston, as well as on the forward camera in Singapore, and the forward-left and forward-right cameras in Boston.
The results can be seen in~\figref{fig:appd_pred_nuscenes} and show that the accuracy is comparable in the four cases.
The sets of intrinsic calibration parameters for the four cameras considered are almost the same and hence much closer in terms of APPD than the ones from the KITTI dataset (less than $10^{-5}$). 
This consistency between the different sensors, as well as the higher resolution of the images, explains why the model trained on nuScenes exhibits better performance.

% \begin{figure}[!t]
% \centering
% \includegraphics[width=0.8\linewidth]{Illustrations/appd_pred_tr_cam2-ts_cam2.pdf}
% \caption{APPD prediction accuracy of the trained neural network model for camera 2 from KITTI \cite{geiger2013vision}.
% %
% Calculation of APPD was performed with respect to both the set of intrinsic calibration parameters corresponding to the day used for training, and for the day corresponding to the image.
% }
% \label{fig:appd_pred_tr_cam2-ts_cam2}
% \end{figure}

% \begin{figure}[!t]
% \centering
% \includegraphics[width=0.8\linewidth]{Illustrations/appd_pred_tr_cam3-ts_cam3.pdf}
% \caption{APPD prediction accuracy of the learnt neural network model for camera 3 from KITTI \cite{geiger2013vision}.
% %
% Calculation of APPD was performed with respect to both the set of intrinsic calibration parameters corresponding to the day used for training, and for the day corresponding to the image.
% }
% \label{fig:appd_pred_tr_cam3-ts_cam3}
% \end{figure}

% \begin{figure}[!t]
% \centering
% \includegraphics[width=0.8\linewidth]{Illustrations/appd_pred_cross_eval.pdf}
% \caption{APPD prediction accuracy of the learnt neural network model for one of the KITTI cameras when evaluated on the data from the other \cite{geiger2013vision}.
% %
% Calculation of APPD was performed with respect to the set of intrinsic calibration parameters of the day corresponding to the image.
% }
% \label{fig:appd_pred_cross_eval}
% \end{figure}

\subsection{Relationship between APPD and Calibration Parameters}
\label{subsec:appd_and_parameters}

Some of the effects of the different intrinsic parameters on the APPD value are illustrated in~\figref{fig:params-to-appd}.
The plots are obtained by individually varying one parameter while keeping the others fixed.
%
%This gives some insights into the metric but does not provide a complete analysis as the influence of the parameters on the APPD are strongly coupled.
%
The difference in effect when varying $f_u$ and $u_c$ compared to $f_v$ and $v_c$, respectively, is due to the wide aspect ratio (2.72) of the image.
This causes parameters along the $u$-axis to have a stronger effect on the distortion of the images.
Another point to observe from~\figref{fig:params-to-appd} is the noise, which is more visible at lower APPD values.
This noise is due to quantization effects causing numerical imprecision when computing the undistortion maps.
The amount of noise can be reduced, at the cost of more computation time, by calculating the APPD at a higher image resolution, but is not necessary for any practical purposes.

\subsection{Relationship between APPD and Reprojection Error}
\label{subsec:appd_and_reprojection}

Reprojection error is a standard measure of the deterioration of a robotic system's performance in various vision-related tasks~\cite{oth2013rolling}.
Therefore, it is of interest to relate the APPD metric of a misrectification to the reprojection error it causes.
As mentioned in \secref{subsec:dataset-generation}, the physical scenario that we are interested in is when a camera experiences a hardware change without the corresponding change in intrinsic parameters.
Data for such scenarios is difficult to obtain.
Therefore we propose applying the reverse process: the physical sensor stays the same while the parameters are changed.
We perform a few simple tests on simulated data to further analyze the relationship between APPD and reprojection error.

\begin{figure}[!t]
\centering
\includegraphics[width=\linewidth]{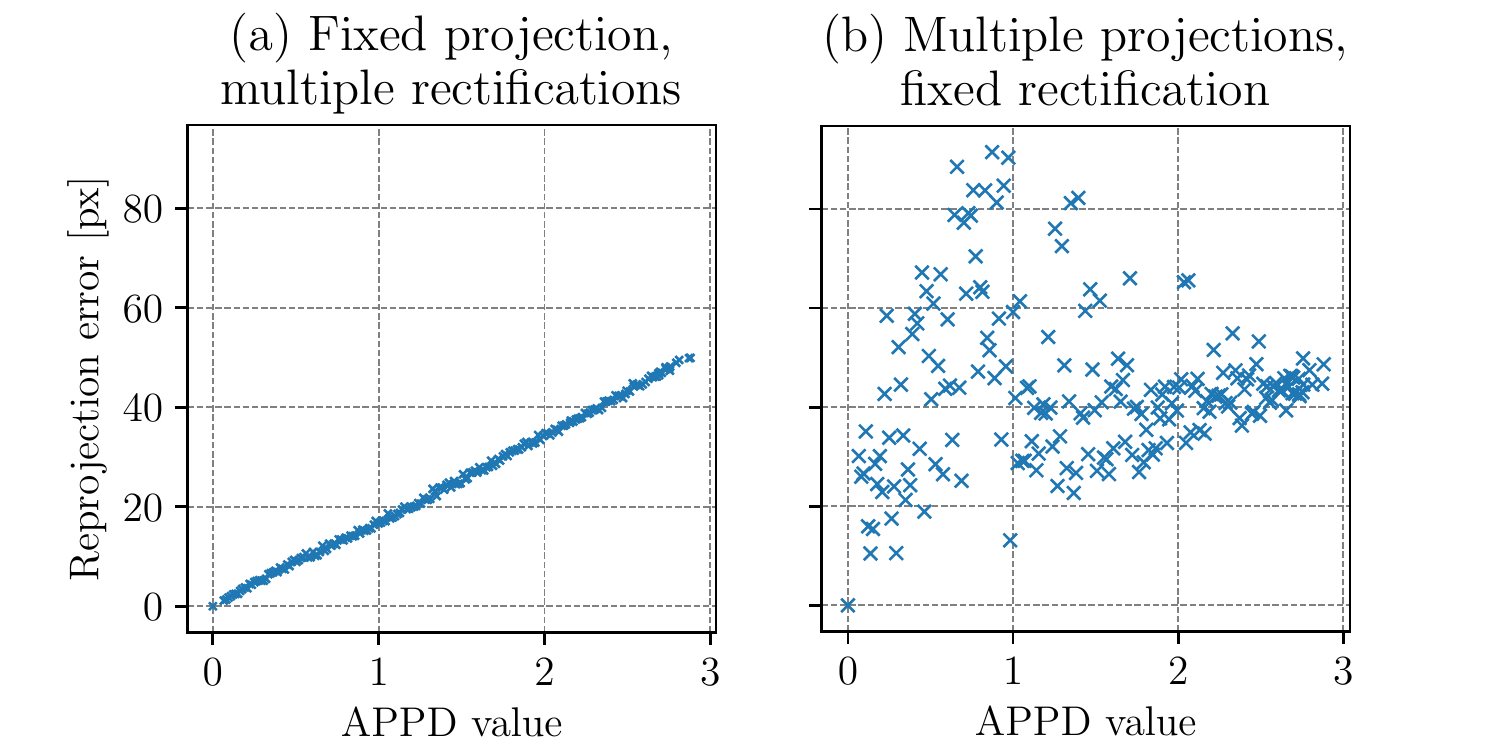}
\caption{Plot between APPD and reprojection error for \textbf{(a)} when projecting a set of points with a single set of calibration parameters and rectifying with a variety of parameters, and \textbf{(b)} when projecting a set of points with a variety of calibration parameters and rectifying with a fixed set.
%
%200 random calibration sets were used for generating the projection parameters.
}
\label{fig:reproj_err_appd}
\end{figure}

First, consider the case when the camera is kept unchanged, but the robot's belief of its intrinsic calibration is changed.
One can generate a set of points in front of a virtual camera and then project them into the camera plane using the correct intrinsic parameters of the physical camera. 
These points can then be rectified with both the correct and incorrect sets of parameters.
Since point associations are known, the reprojection error can be calculated as the average distance between the resulting rectified projections in the image plane.
\mbox{\figref{fig:reproj_err_appd}a} shows the obtained relationship.
Indeed, APPD is a good measure of the reprojection error that arises from rectifying with a wrong calibration parameter set.

Second, it is of interest to know how the real physical scenario would relate with the reverse synthetic scenario in order to evaluate if the method outlined here can be applied to a real system.
This can be achieved by repeating the above-described point-projection and rectification procedure, but keeping the intrinsic parameters for the rectification step fixed while varying the set for the projection step.
This setting corresponds exactly to the physical situation but cannot be reproduced synthetically on a real image (we cannot `reproject' reality with a different set of intrinsic parameters).
The comparison between the resulting reprojection error and APPD can be seen in~\mbox{\figref{fig:reproj_err_appd}b}.
The result is that there is no-longer an injective functional relationship from APPD to reprojection error, and the dependence between the two values is less pronounced. 

%
%%% THE PLOTS ARE CHANGED AND WHILE THESE STATEMENTS ARE TRUE, WE CANNOT CLAIM THEM UNLESS WE PUT ANOTHER PLOT, SOMETHING I THINK IS UNNECESSARY ~Aleks
% Second, for each APPD value, we observe a pair of reprojection error values. 
%
% These correspond to the cases when the parameter set used for projection and the one for rectification are switched.
%
% This is an expected result as while APPD is commutative, the reprojection error is not, i.e. switching the projection and rectification parameters generally results in different average reprojection error.

\begin{figure}[!t]
\centering
\includegraphics[width=\linewidth]{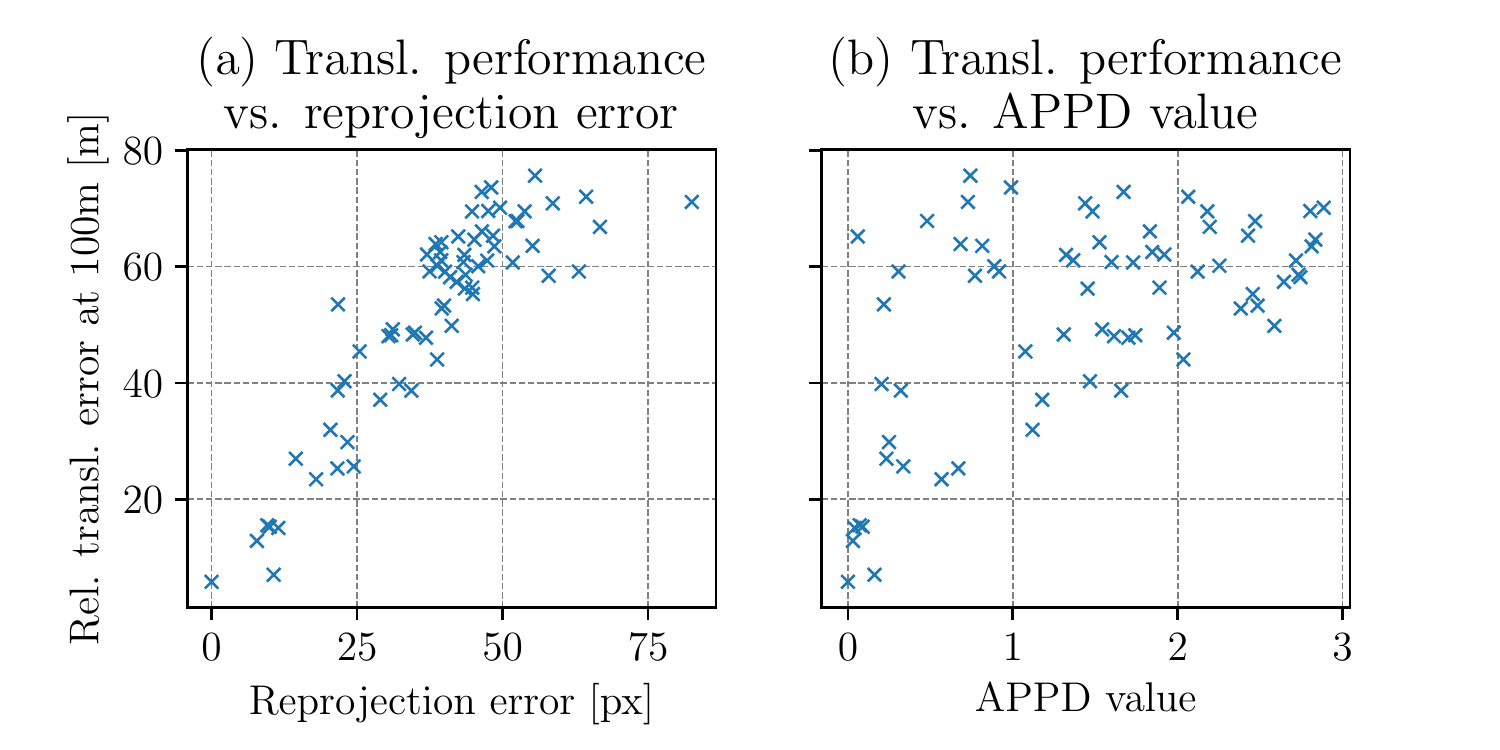}
\caption{
Plot between the performance of ORB-SLAM~\cite{mur2017orb}, evaluated on the KITTI odometry sequence 10, and \textbf{(a)} reprojection error when projecting a set of points with a single set of parameters and rectifying with a variety of sets, and \textbf{(b)} the corresponding APPD.
}
\label{fig:reproj_err_ORBSLAM_performance}
\end{figure}

APPD is easy to calculate for a real-world dataset as it is independent of the hardware that is used to obtain the image.
Furthermore, it allows the sampling of an almost infinite number of different intrinsic calibration parameters, which is beneficial for training neural networks, which require large amounts of data. 
Nevertheless, it might not be the most accurate metric for detecting physical miscalibration. 
In fact, as~\figref{fig:reproj_err_ORBSLAM_performance} shows, the reprojection error as computed for the physical scenario is better correlated with the SLAM performance of a system.
As mentioned above, the drawback of using reprojection error as a miscalibration metric is that it cannot be calculated for a real-world dataset.
No procedure similar to the one in \secref{subsec:dataset-generation} can be constructed for the physical miscalibration case and its corresponding reprojection error.
Therefore, one would need to create a dataset with various camera settings and the respective calibration parameters for each one, which can be impractically time-consuming as it needs to be repeated for each camera individually.
The variety of possible calibrations would also be severely limited by the design of the lens, as most lenses only have one degree of freedom.
Alternatively, a fully synthetic dataset, e.g. generated in simulation, can be used, but then transferability to real image data would be questionable.

The advantage of using APPD is that it facilitates training with very large sets of data that are easily obtained via the procedure detailed in~ \secref{subsec:dataset-generation}. 
The type of artifacts introduced by the dataset generation in~\secref{subsec:dataset-generation} can be considered similar to the ones introduced by a physical miscalibration. 
By demonstrating that APPD is learnable by a neural network, we show that it might be possible to also learn the reprojection error.

%\subsection{How to choose an APPD value?}
%
%We show in \figref{fig:} the performance of ORB-SLAM~\cite{mur2017orb}, a monocular visual-odometry algorithm, as a function of APPD.
%
%The evaluations are done on sequence \texttt{2011\_09\_30\_drive\_0034\_camera0} from the KITTI dataset \cite{geiger2013vision}, with both the correct and incorrect calibrations.
%
%Based on \figref{fig:} we can choose $\delta_{\text{thresh}} = \red{XX}$ as a reasonable operating point, to decide when a miscalibration has occurred.

\section{Conclusion}

We proposed a novel semi-synthetic data generation procedure that requires no data labeling and a corresponding camera miscalibration metric called the \textit{average pixel position difference (APPD)}.
These tools can then be used to train a simple CNN, which we show is able to predict the APPD values from images with no additional data necessary.
The performance of the network was evaluated on different real-world datasets and cameras.
Provided the camera's true intrinsic parameters remained close, the network was able to generalize well to different cameras and environments that it had not seen before.
Such a network can then be deployed on a real robotic platform, running at a very low frequency, to determine if a more expensive recalibration procedure needs to be executed.

\addtolength{\textheight}{-5.5cm}   % This command serves to balance the column lengths

%%%%%%%%%%%%%%%%%%%%%%%%%%%%%%%%%%%%%%%%%%%%%%%%%%%%%%%%%%%%%%%%%%%%%%%%%%%%%%%%

%%%%%%%%%%%%%%%%%%%%%%%%%%%%%%%%%%%%%%%%%%%%%%%%%%%%%%%%%%%%%%%%%%%%%%%%%%%%%%%%



\section*{ACKNOWLEDGMENTS}
The authors would like to thank Jen Jen Chung, Lionel Ott, Juan Nieto and Davide Scaramuzza for their feedback and valuable insights.

%%%%%%%%%%%%%%%%%%%%%%%%%%%%%%%%%%%%%%%%%%%%%%%%%%%%%%%%%%%%%%%%%%%%%%%%%%%%%%%%

{\small
\bibliographystyle{IEEEtran}
\bibliography{main}
}

\end{document}